\documentclass[]{article}
\usepackage{booktabs}
\usepackage{stackrel}
\usepackage{proceed2e}
\usepackage{times} % Use times fonts.
\usepackage{subscript} % A robust text-mode subscript
\usepackage[geometry]{ifsym} % Circles, squares, diamonds...
\usepackage{color} % Colored text
\usepackage[utf8]{inputenc} % Type accents, ligatures etc directly.
\usepackage{amsmath,amssymb,amsthm} % Math fonts and environments
\usepackage{graphicx} % Reasonable graphic formats
\usepackage{subfigure} % Subfigures with sublabels 
\usepackage{booktabs} % Not-horrible tables
\usepackage{multirow} % Multi-row table functionality
\usepackage{multicol}
\usepackage[dvipsnames]{xcolor}
\usepackage[round]{natbib}
\usepackage{url}
\usepackage{caption}
\usepackage{paralist}
\usepackage[lined, ruled, boxed, linesnumbered, commentsnumbered]{algorithm2e}
\usepackage{titlesec} % Modify spacing around various headers
\titlespacing\section{0pt}{12pt plus 4pt minus 2pt}{0pt plus 2pt minus 2pt}
\titlespacing\subsection{0pt}{12pt plus 4pt minus 2pt}{0pt plus 2pt minus 2pt}
\titlespacing\subsubsection{0pt}{12pt plus 4pt minus 2pt}{0pt plus 2pt minus 2pt}

\newcommand{\calf}{\mathcal{F}}

\usepackage{centernot}
\newcommand{\CI}{\mathrel{\perp\mspace{-10mu}\perp}}

\newcommand{\ef}{ef\!fect}
\newcommand{\cs}{cause}
\newcommand{\pxy}{P_{XY}}
\newcommand{\px}{P_X}
\newcommand{\pymidx}{P_{Y\mid X}}

\usepackage{times}

\title{Estimating Causal Direction and Confounding Of Two Discrete Variables}
\author{ {\bf Krzysztof Chalupka} \\
Computation and\\ Neural Systems\\
Caltech\\
\And
{\bf Frederick Eberhardt}   \\
Humanities and\\ Social Sciences\\
Caltech\\
\And
{\bf Pietro Perona}  \\
Electrical Engineering\\
Caltech\\
}
\begin{document}
\maketitle

\begin{abstract}
We propose a method to classify the causal relationship between two discrete variables given only the joint distribution of the variables, acknowledging that the method is subject to an inherent baseline error. We assume that the causal system is acyclicity, but we do allow for hidden common causes. Our algorithm presupposes that the probability distributions $P(C)$ of a cause $C$ is independent from the probability distribution $P(E\mid C)$ of the cause-effect mechanism. While our classifier is trained with a Bayesian assumption of flat hyperpriors, we do not make this assumption about our test data. This work connects to recent developments on the identifiability of causal models over continuous variables under the assumption of "independent mechanisms". Carefully-commented Python notebooks that reproduce all our experiments are available online at \url{vision.caltech.edu/~kchalupk/code.html}.

%It is not possible in general to identify the causal relationship between two discrete variables given only the joint distribution of the variables. We propose a method to classify the causal relationship nevertheless, acknowledging that the method is subject to an inherent baseline error. We assume acyclicity of the causal system, but we do allow for hidden common causes. Our algorithm presupposes that the probability distributions of causes are independent from probability distributions of cause-effect mechanisms. In addition, we rely on weak Bayesian assumptions (flat hyperpriors). Well-commented Python notebooks that reproduce all our experiments are available online at \url{vision.caltech.edu/~kchalupk/code.html}.
\end{abstract}

\section{Introduction}
Take two discrete variables $X$ and $Y$ that are probabilistically dependent. Assume there is no feedback between the variables: it is not the case that both $X$ causes $Y$ and $Y$ causes $X$. Further, assume  that any probabilistic dependence between variables  always arises due to a causal connection between the variables \citep{Reichenbach1991}. The fundamental causal question is then to answer two questions: 1) Does $X$ cause $Y$ or does $Y$ cause $X$? And 2) Do $X$ and $Y$ have a common cause $H$? Since we assumed no feedback in the system, the options in 1) are mutually exclusive. Each of them, however, can occur together with a confounder. Fig.~\ref{fig:causal_alternatives} enumerates the set of six possible hypotheses. 

In this article we present a method to distinguish between these six hypotheses on the basis only of data from the observational joint probability $P(X,Y)$ -- that is, without resorting to experimental intervention.

Within the causal graphical models framework~\citep{Pearl2000,Spirtes2000}, differentiating between any two of the causally interesting possibilities (shown in Fig.~\ref{fig:causal_alternatives}B-F) is in general only possible if one has the ability to intervene on the system. For example, to differentiate between the pure-confounding and the direct-causal case (Fig.~\ref{fig:causal_alternatives}B and C), one can intervene on $X$ and observe whether that has an effect on the distribution of $Y$. Given only observations of $X$ and $Y$ and no ability to intervene on the system however, the problem is in general not identifiable. %Specifically, for causal models with linear Gaussian or multinomial parameterizations, \citet{gp1988} and \citet{Meek1995}, respectively, showed that any method that identifies the Markov equivalence class of causal models, is complete. Hypotheses \ref{fig:causal_alternatives}B-F fall in the same Markov equivalence class. For the multinomial case, the reason is intuitive. 
Roughly speaking, the reason is simply that without any further assumptions about the form of the distribution, any joint $P(X,Y)$ can be factorized as $P(X)P(Y\mid X)$ and $P(Y)P(X\mid Y)$, and the hidden confounder $H$ can easily be endowed with a distribution that can give the marginal $\sum_H P(X,Y,H)$ any desired form.

\section{Related Work}
\label{sec:related}
There are two common remedies to the fundamental unidentifiability of the two-variable causal system: 1) Resort to interventions or 2) Introduce additional assumptions about the system and derive a solution that works under these assumptions.

Whereas the first solution is straightforward, research in the second direction is a more recent and exciting enterprise. 
\subsection{Additive Noise Models}
A recent body of work attacks the problem of establishing whether $x\rightarrow y$ or $y \rightarrow x$ when specific assumptions with respect to the functional form of the causal relationship hold. \citet{Shimizu2006} showed for continuous variables that when the effect is a linear function of the cause, with \emph{non-Gaussian} noise, then the causal direction can be identified in the limit of infinite sample size.

This inspired further work on the so called ``additive noise models''. \citet{Hoyer2009} extended Shimizu's identifiability results to the case when the effect is any (except for a measure-theoretically small set of) nonlinear function of the cause, and the noise is additive -- even Gaussian. \citet{Zhang2009} showed that a \emph{post}nonlinear model -- that is, $y = f(g(x) + \epsilon)$ where $f$ is an invertible function and $\epsilon$ a noise term -- is identifiable. The additive noise models framework was applied to discrete variables by~\citet{Peters2011}. 

Most of this work has focused on  distinguishing the causal orientation in the absence of confounding (i.e.\ distinguishing hypotheses A, C and D in Fig.~\ref{fig:causal_alternatives}, although \cite{hskp2008} have extended the linear non-Gaussian methods to the general hypothesis space of Fig,~\ref{fig:causal_alternatives} and \citet{Janzing2009} showed that the additive noise assumption can be used to detect pure confounding with some success, i.e.\ to distinguish hypothesis B from hypotheses C and D.

The assumption of additive noise supplies remarkable identifiability results, and has a natural application in many cases where the variation in the data is thought to derive from the measurement process on an otherwise deterministic functional relation. With respect to the more general space of possible probabilistic causal relations it constitutes a very substantive assumption. In particular, in the discrete case its application is no longer so natural.  

\subsection{Bayesian Causal Model Selection}
From a Bayesian perspective, the question of causal model identification is softened to the question of model selection based on the posterior probability of a causal structure give the data. In a classic work on Bayesian Network learning (then called Belief Net learning), Heckerman and Chickering~\citep{Heckerman1995,Chickering2002,Chickering2002b} developed a Bayesian scoring criterion that allowed them to define the posterior probability of each possible Bayesian network given a dataset. 
Motivated by results of \citet{gp1988} and \citet{Meek1995} that showed that for linear Gaussian parameterizations and for multinomial parameterizations, causal models with identical (conditional) independence structure cannot in principle be distinguished given only the observational joint distribution over the observed variables, their score had the property that it assigned the same score to graphs with the same independence structure. But the results of \citet{gp1988} and \citet{Meek1995} only make an existential claim: That for any joint distribution a parameterization of the appropriate form exists for any causal structure in the Markov equivalence class. But one need not conclude from such an existential claim, like Heckerman and Chickering did,  that there aren't reasons in the data that could suggest that one  causal structure in a Markov equivalence class is more probable than another.

%This work introduces five assumptions that together define which networks are more and less likely. Their Assumptions 1 (Multinomial Probabilities), 2 (Parameter Independence) and 5 (Multinomial Hyperpriors) can be used to define the likelihood of the structures shown here in Fig.~\ref{fig:causal_alternatives}. We do not repeat the assumptions here, as we propose their modified versions in Sec.~\ref{sec:assumptions}.

\begin{figure}
\centering
\includegraphics[width=.4\textwidth]{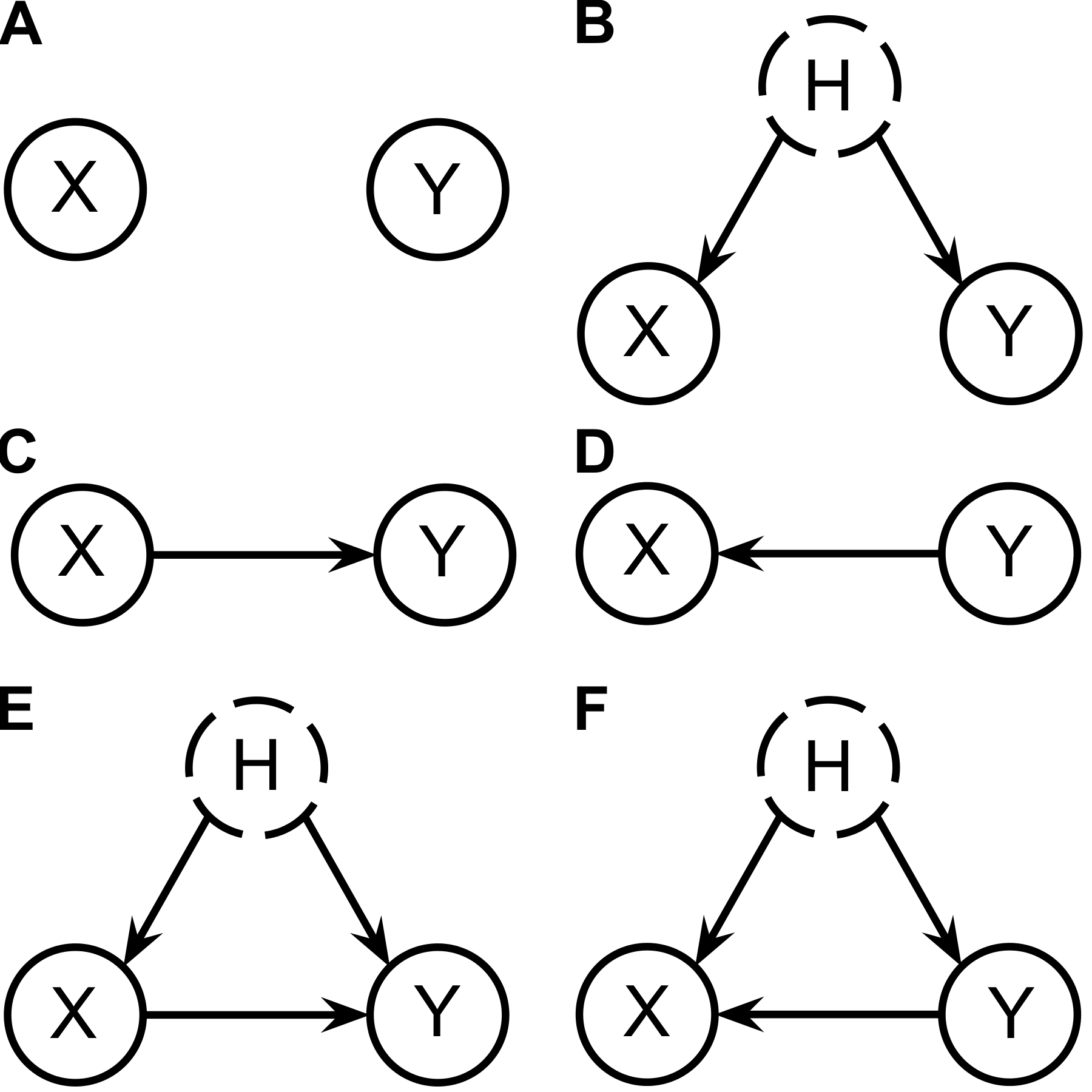}
\caption{Possible causal structures linking X and Y. Assume X, Y and H are all discrete but H is unobserved. In principle, it is impossible to identify the correct causal structure given only X and Y samples. In this report, we will tackle this problem using a minimalistic set of assumptions. Our final result is a classifier that differentiates between these six cases -- the confusion matrices are shown in Fig.~\ref{fig:all_results}.}
\label{fig:causal_alternatives}
\end{figure}

Thus, the crucial difference between the work of Heckermann and ours is that their goal was to find the \emph{Markov Equivalence Class} of the true causal model. 
%That is, to them two networks that encode the same independence assumptions are equivalent. 
This, however, renders our task impossible: all the hypotheses enumerated in Fig.~\ref{fig:causal_alternatives}B-F are Markov-equivalent. Our contribution is to use assumptions formally identical to theirs (when restricted to the causally sufficient case) to assess the likelihood of causal structures over two observed variables, bearing in mind that even in the limit of infinitely many samples, the true model cannot be determined, the structures can only be deemed more or less likely.

%there is no ``right'' structure for any observed joint, but there are ``more likely'' and ``less likely'' structures.

Our approach is most similar in spirit to the work of~\citet{Sun2006}. Sun puts an explicit Bayesian prior on what a likely causal system is: if $X$ causes $Y$, then the conditionals $p(Y\mid X=x)$ are \emph{less complex} than the reverse conditionals $P(X \mid Y=y)$, where complexity is measured by the Hilbert space norm of the conditional density functions. This formulation is plausible and easily applicable to discrete systems (by defining the complexity of discrete probability tables by their entropy).

\section{Assumptions}
\label{sec:assumptions}

%\subsection{Desiderata}
 Our contribution is to create an algorithm with the following properties:
\begin{compactenum}
\item Applicable to \emph{discrete} variables $X$ and $Y$ with finitely many states.
\item Decides between all the six possible graphs shown in Fig.~\ref{fig:causal_alternatives}.
\item Does not make assumptions about the functional form of the discrete parameterization (in contrast to e.g.\ an additive noise assumption).
\end{compactenum}

In a recent review,~\citet{Mooij2014} compares a range of methods that decide the causal direction between two variables, including the methods discussed above. To our knowledge, none of these methods attempt to distinguish between the pure-causal (A, C, D), the confounded (B), and the causal+confounded case (E,F).

We take an approach inspired by the Bayesian methods discussed in Sec.~\ref{sec:related}. Consider the Bayesian model in which $P(X,Y)$ is sampled from an uninformative hyperprior with the property that the distribution of the cause is independent of the distribution of the effect conditioned on the cause:

\begin{compactenum}
\item Assume that $P(\ef \mid \cs) \CI P(\cs)$. 
\item Assume that $P(\ef \mid \cs =c)$ is sampled from the uninformative hyperprior for each $c$.
\item Assume that $P(\cs)$ is sampled from the uninformative hyperprior.
\end{compactenum}

Since all the distributions under considerations are multinomial, the ``uninformative hyperprior'' is the Dirichlet distribution with parameters all equal to 1 (which we will denote as $Dir(1)$, remembering that $1$ is actually a vector whose dimensionality will be clear from context). Which variable is a cause, and which the effect, or whether there is confounding, depends on which of the causal systems in Fig.~\ref{fig:causal_alternatives} are sampled.
For example, if $X \rightarrow Y$ and there is also confounding $X \leftarrow h \rightarrow Y$ (Fig.~\ref{fig:causal_alternatives}D), then our assumptions set
\begin{align*}
P(X) &\sim Dir(1)\\
\forall_x P(Y\mid X=x) &\sim Dir(1)\\
P(H) &\sim Dir(1)\\\
\forall_h P(X \mid H=h) &\sim Dir(1)\\
\forall_h P(Y \mid H=h) &\sim Dir(1)
\end{align*}

\section{An Analytical Solution: Causal Direction}
\label{sec:analytical}
Consider first the problem of identifying the causal direction. That is, assume that either $X \rightarrow Y$ or $Y \rightarrow X$, and there is no confounding. The assumptions of Sec.~\ref{sec:assumptions} then allow us to compute, for any given joint $P(X, Y)$ (which we will from now on denote $\pxy$ to simplify notation), the likelihood $p(X\rightarrow Y \mid \pxy)$ and the likelihood $p(Y\rightarrow X \mid \pxy)$. The likelihood ratio allows us to decide which causal direction $\pxy$ more likely represents. 

We first derive and visualize the likelihood for the case of $X$ and $Y$ both binary variables. Next, we generalize the result to general $X$ and $Y$. Finally, we analyze experimentally how sensitive such causal direction classifier is to breaking the assumption of uninformative Dirichlet hyperpriors (but keeping the independent mechanisms assumption).

\subsection{Optimal Classifier for Binary $X$ and $Y$}

Consider first the binary case. Let $\px = \begin{bmatrix}a\\ 1-a\end{bmatrix}$ and $\pymidx = \begin{bmatrix}b & 1-b\\ c & 1-c\end{bmatrix}$. Assume $\px$ is sampled independently from $\pymidx$, and that the densities (parameterized by $a$ and $b,c$) are $\mathcal{F}_a, \mathcal{F}_b, \mathcal{F}_c\colon (0, 1) \rightarrow \mathbb{R}$. This defines a density over $(a, b, c)$, the three-dimensional parameterization of an $x \rightarrow y$ system, as $\mathcal{F}(a, b, c) = \calf_a(a)\calf_b(b)\calf_c(c) \colon (0,1)^3 \rightarrow \mathbb{R}$.

Now, consider $\pxy = \begin{bmatrix}d & e\\ f & 1-(d+e+f)\end{bmatrix}$ -- a three-dimensional parameterization of the joint. If we assume that $\pxy$ is sampled according to the $X \rightarrow Y$ sampling procedure, we can compute its density $\mathcal{H}_{XY}\colon (0,1)^3 \rightarrow \mathbb{R}$ as a function of $\calf$ using the multivariate change of variables formula. We have 
\[
\begin{bmatrix}d\\ e\\ f \end{bmatrix} = 
\begin{bmatrix}ab\\ a(1-b)\\ (1-a)c\end{bmatrix}
\]
and the inverse transformation is
\begin{equation}
\begin{bmatrix}a\\ b\\ c \end{bmatrix} = 
\begin{bmatrix}d+e\\\frac{d}{d+e}\\\frac{f}{1-d-e}\end{bmatrix}\label{eq:inverse_joint}
\end{equation}
The Jacobian of the inverse transformation is
\[
\frac{d(a,b,c)}{d(d,e,f)} =
\begin{bmatrix}
1 & 1 & 0\\
\frac{e}{(d+e)^2} & \frac{-d}{(d+e)^2} & 0\\
\frac{f}{(1-d-e)^2} & \frac{f}{(1-d-e)^2} & \frac{1}{1-d-e}, 
\end{bmatrix}
\]
its determinant $\det\left(\frac{d(a,b,c)}{d(d,e,f)}\right) = \frac{-1}{(d+e)-(d+e)^2}$. The change of variables formula then gives us
\begin{equation*}
\mathcal{H}_{XY}(d,e,f) = \frac{\calf(d+e, \frac{d}{d+e},\frac{f}{1-d-e})}{(d+e)-(d+e)^2},
\end{equation*}
where $a,b,c$ are obtained from Eq.~\eqref{eq:inverse_joint}.

We can repeat the same reasoning for the inverse causal direction, $Y\rightarrow X$. In this case, we obtain

\begin{equation*}
\mathcal{H}_{YX}(d,e,f) = \frac{\calf(d+f, \frac{d}{d+f}, \frac{e}{1-d-f})}{(d+f)-(d+f)^2}.
\end{equation*}

Given $\pxy$ and the hyperpriors $\calf$, we can now test which causal direction $\pxy$ most likely corresponds to. Assuming equal priors on both causal directions, we have

\begin{align*}
&\frac{p(X\rightarrow Y \mid (d,e,f))}{p(Y\rightarrow X \mid (d,e,f))} =  \frac{\mathcal{H}_{xy}(d,e,f)}{\mathcal{H}_{yx}(d,e,f)}\\
&=  \frac{\calf\left(d+e, \frac{d}{d+e}, \frac{f}{1-d-e}\right)}{\calf\left(d+f, \frac{d}{d+f}, \frac{e}{1-d-f}\right)}\frac{(d+f)-(d+f)^2}{(d+e)-(d+e)^2}
\end{align*}

\begin{figure}
\includegraphics[width=.5\textwidth]{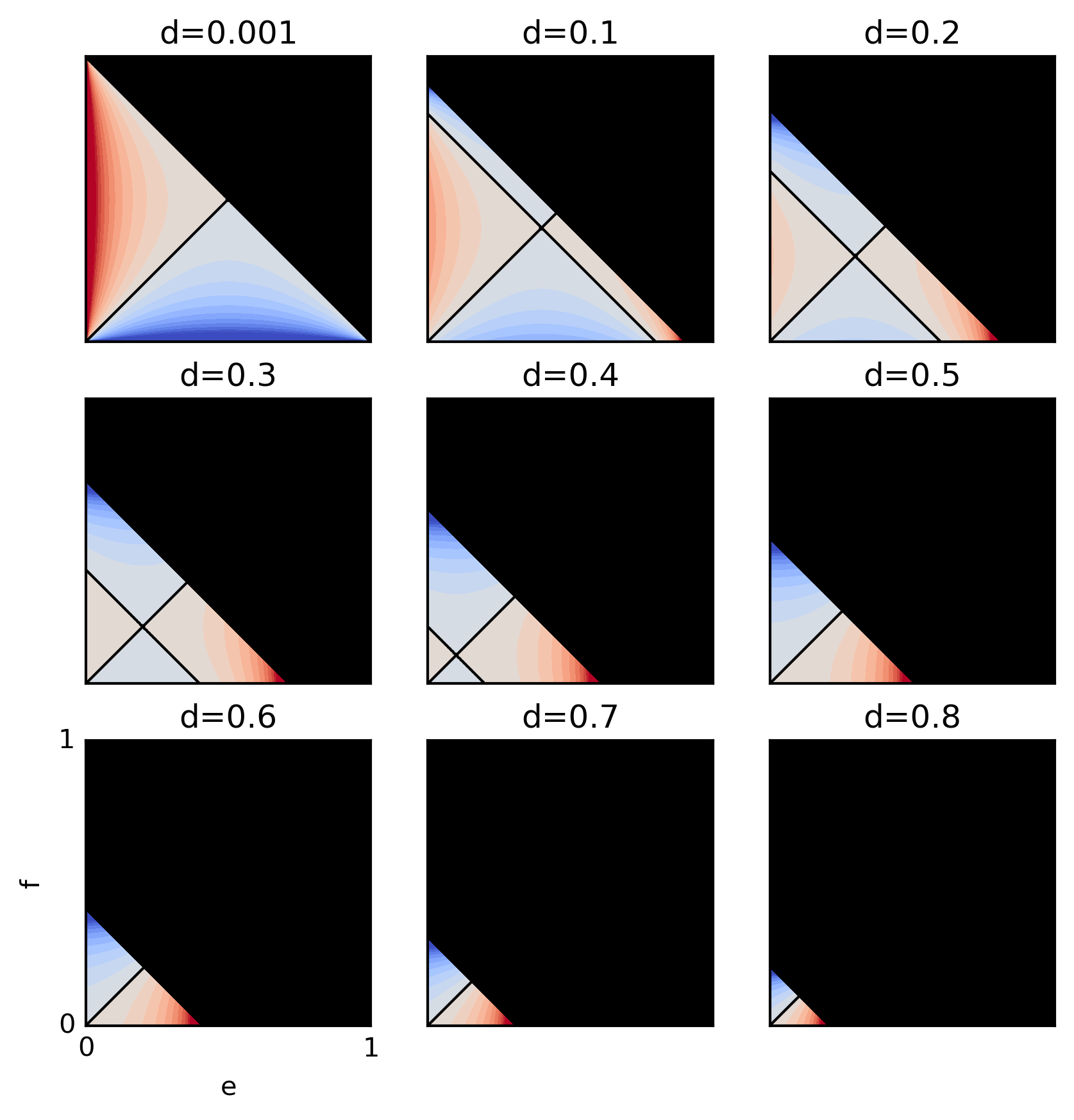}
\caption{Log likelihood-ratio $\log\left(\frac{P(X\rightarrow Y \mid (d,e,f))}{P(Y\rightarrow X \mid (d,e,f))}\right)$ as a function of $e,f$ for nine different values of $d$. Red corresponds to values larger than 0 --- that is, $X\rightarrow Y$ is more likely than the opposite causal direction in the red regions. Blue signifies the opposite. The decision boundary is shown in black. It is a union of two orthogonal planes that cut the $(d,e,f)$ simplex into four connected components along a skewed axis. }
\label{fig:loglik}
\end{figure}

Only the first factor in the likelihood ratio depends on the hyperprior $\calf$. If we fix $\calf_a,\calf_b,\calf_c$ to all be $Dir(1)$, the factor reduces to 1 and the likelihood ratio becomes

\begin{equation*}
\frac{p(X\rightarrow Y \mid (d,e,f))}{p(Y\rightarrow X \mid (d,e,f))} =  \frac{(d+f)-(d+f)^2}{(d+e)-(d+e)^2}.
\end{equation*}

Denote the ``uninformative-hyperprior likelihood ratio'' function \[LR \colon \pxy(d,e,f) \mapsto \frac{(d+f)-(d+f)^2}{(d+e)-(d+e)^2}.\]
The classifier that assigns the $X\rightarrow Y$ class to $\pxy$ with $LR(\pxy) > 1$, and the $Y \rightarrow X$ class otherwise is the optimal classifier under our assumptions. Fig.\ref{fig:loglik} shows LR across the three-dimensional $\pxy$ simplex. The figure shows nine slices of this simplex for different values of the $d$ coordinate. 

\subsection{Optimal Classifier for Arbitrary $X$ and $Y$}
Deriving the optimal classifier for the case where $X$ and $Y$ are not binary is analogous to the binary derivation. The resulting likelihood ratio is 

\begin{align}
&\frac{p(X\rightarrow Y \mid \pxy)}{p(Y\rightarrow X \mid \pxy)} = \\
&=  \frac{\calf\left(P_X, P_{Y\mid X}\right)}{\calf\left(P_Y, P_{X\mid Y}\right)}\frac{|\det{J_{XY}}|^{-1}}{|\det{J_{YX}}|^{-1}},\label{eq:lr_general}
\end{align}

where $J_{XY}$ is the Jacobian of the linear transformation $(P_X, P_{Y\mid X}) \mapsto P_{XY}$ and $J_{YX}$ is the Jacobian of the transformation $(P_Y, P_{X\mid Y}) \mapsto P_{XY}$. The transformation, its determinant and Jacobian are readily computable on paper or using computer algebra systems. In our implementation, we used Theano~\citep{Theano} to perform the computation for us. Note that if $X$ has cardinality $k_X$ and $Y$ has cardinality $k_Y$, the Jacobians have $(k_Xk_Y - 1)^2$ entries. Computing their determinants has complexity $\mathcal{O}((k_Xk_Y-1)^6)$ or, if we assume $k_X = k_Y = k$, $\mathcal{O}(k^{12})$ -- it grows rather quickly with growing cardinality.

If $\calf$ is flat, that is all the priors are $Dir(1)$, we will call the causal direction classifier that follows Eq.~\eqref{eq:lr_general} the LR classifier. That is, the LR classifier outputs $X\rightarrow Y$ if the uninformative-hyperprior likelihood ratio is larger than 1, and outputs $Y\rightarrow X$ otherwise.

Note that the \emph{optimal} classifier is not \emph{perfect} -- there is a baseline error that the optimal classifier has under the assumptions it is built on. This error is
\begin{align*}
E_{LR} &= \int p(Y\rightarrow X \mid \pxy) I_{[LR(\pxy) > 1]}+\\
&\phantom{W} p(X\rightarrow Y\mid \pxy) I_{[LR(\pxy) < 1]} d\pxy,
\end{align*}
where the integral varies over all the possible joints $\pxy$ with uniform measure, and $I_{[LR(\pxy) <> 1]}$ is the indicator function that evaluates to 1 if its subscript condition holds, and to 0 otherwise.

That is, assuming that each $\pxy$ is sampled from the uninformative Dirichlet prior given that either $X\rightarrow Y$ or $Y\rightarrow X$ with given probability, in the limit of infinite classification trials the error rate of the LR classifier is $E_{LR}$. Whereas this integral is not analytically computable (at least neither by the authors nor by available computer algebra systems), we can estimate it using Monte Carlo methods in the following sections. In Fig.~\ref{fig:dmm_results}, the leftmost entry on each curve corresponds to $E_{LR}$ for various cardinalities of $X$ and $Y$. For example, for $|X|=|Y|=2$, $E_{LR} \approx .4$ but already for $|X|=|Y|=10$, $E_{LR} < .001$.

\subsection{Robustness: Changing the Hyperprior $\calf$}
What if we use the LR classifier, but our assumptions do not match reality? Namely, what if $\calf$ is \emph{not} $Dir(1)$? For example, what if $\calf$ is a mixture of ten Dirichlet distributions\footnote{A mixture of Dirichlet distributions with arbitrary many components can approximate any distribution over the simplex.}? 

\begin{figure}
\centering
\includegraphics[width=.5\textwidth]{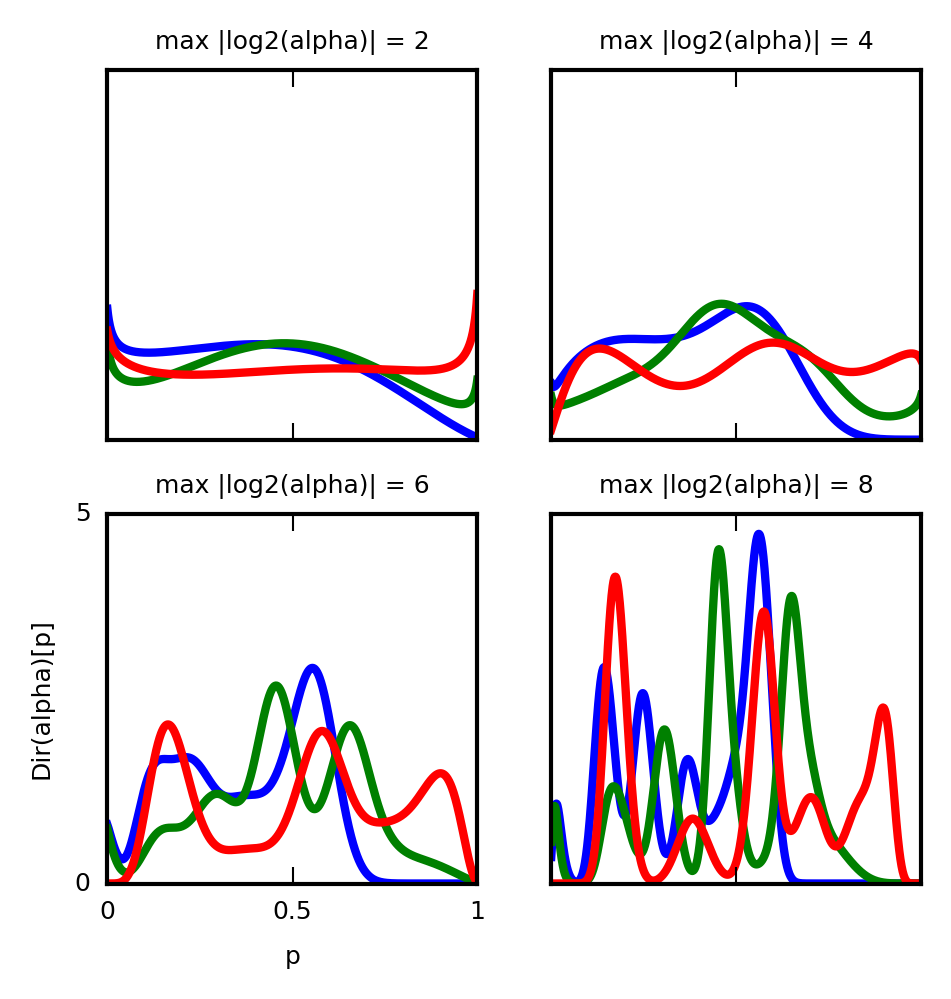}
\caption{Samples from Dirichlet mixtures. Each plot shows three random samples from a ten-component mixture of Dirichlet distributions over the 1D simplex. Each mixture component has a different, random parameter $\alpha$. For each plot we fixed a different $|log_2(\alpha_{max})|$, a parameter which limits both the smallest and largest value of any of the two $\alpha$ coordinates that define each mixture component.}
\label{fig:mixtures}
\end{figure}

\begin{figure}
\centering
\includegraphics[width=.5\textwidth]{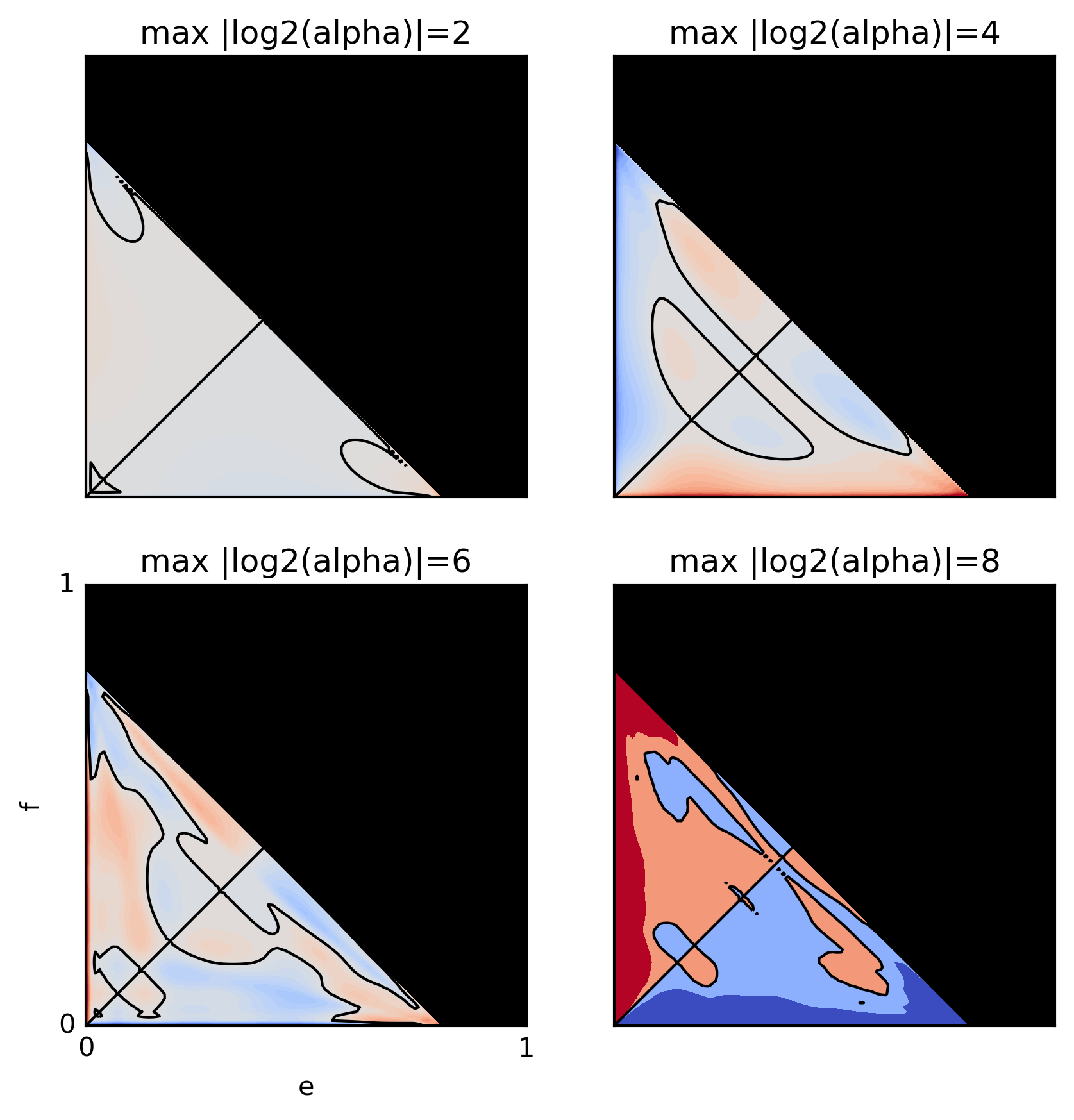}
\caption{Log-likelihood ratios for the causal direction when $\calf$ is a mixture of ten Dirichlet distributions with growing $\alpha_{max}$ (see Fig.~\ref{fig:mixtures}).}
\label{fig:mixture_logliks}
\end{figure}

We will draw $\calf$ from mixtures with fixed ``$|\log_2(\alpha_{max})|$''. Let the $k$-th component of the mixture have parameter $\alpha^k = (\alpha^k_1, \cdots, \alpha^k_N)$ where $N$ is the cardinality of $X$ or $Y$. Then fixed $\alpha_{max}$ means that we drew each $\alpha^k_i$ uniformly at random from the interval $2^{-\alpha_{max}}, 2^{\alpha_{max}}$.
Fig.~\ref{fig:mixture_logliks} shows samples from such mixtures with growing $\alpha_{max}$. The figure shows that increasing the parameter allows the distributions to grow in complexity. 

\begin{figure}
\centering
\includegraphics[width=.5\textwidth]{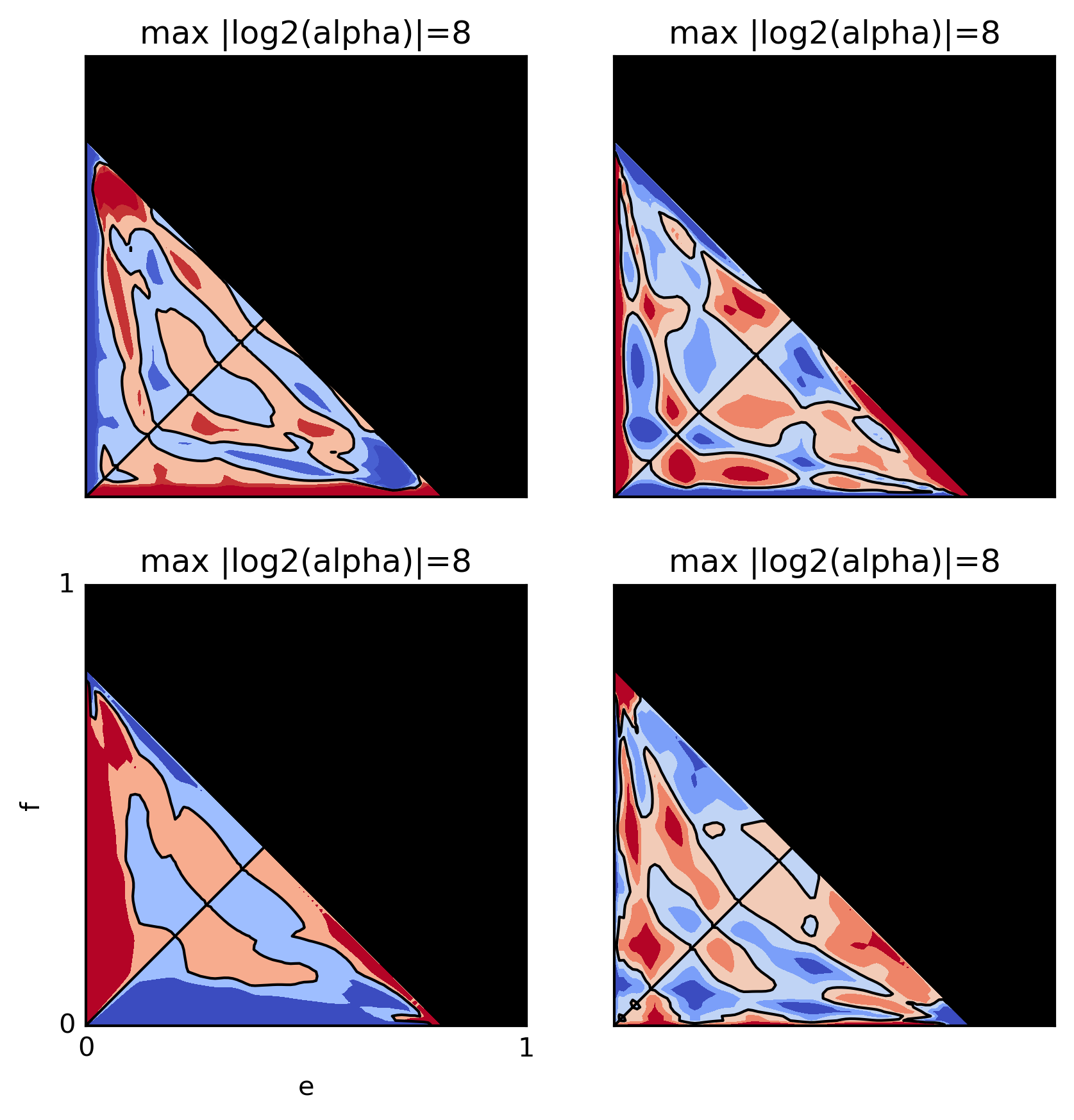}
\caption{Log-likelihood ratios for the causal direction when $\calf$ is a mixture of ten Dirichlet distributions with $|\alpha_{max}| = 2^8$ (see Fig.~\ref{fig:mixtures}) -- each plot corresponds to different, randomly sampled $\alpha$.}
\label{fig:mixture_logliks_alpha8}
\end{figure}

Note that if $\alpha_{max} = 0$, we recover the noninformative prior case. How does the likelihood ratio and the causal direction decision boundary change as we allow $\alpha_{max}$ to depart from $0$? For binary $X$ and $Y$, Fig.~\ref{fig:mixture_logliks} illustrates the change. Comparing with Fig.~\ref{fig:loglik}, we see that as $\alpha_{max}$ grows, the likelihood ratios become more extreme, and the decision boundaries become more complex. Fig.~\ref{fig:mixture_logliks_alpha8} makes it clear that a fixed $\alpha_{max}$ allows for the decision boundary to vary significantly.

That the ``independent mechanisms'' assumption as we framed it is not sufficient to provide identifiability of the causal direction was clear from the outset (since each joint can be factorized as $P(X)P(Y\mid X)$ and $P(Y)P(X\mid Y)$). However, the above considerations suggest that the assumption of noninformative hyperpriors is rather strong: In fact, it is possible to show that the decision surface can be precisely flipped with appropriate adjustment of $\calf$, making the $LR$ classifier's error precisely $100\%$.

Our experiments, however, suggest that using the $LR$ classifier is a reasonable choice in a wide range of circumstances, \emph{especially as the cardinality of $X$ and $Y$ grows}. In our experiments, we checked how the error changes as we allow the $\alpha_{max}$ parameter of all the hyperpriors to grow. Our experimental procedure is as follows:

\begin{compactenum}
\item \textbf{Fix the dimensionality} of $X$ and $Y$, and fix $\alpha_{max}$.
\item \textbf{Sample 100 hyperpriors for each dimensionality and $\alpha_{max}$}. Sample $\alpha$ parameters for $\calf$ within given $\alpha_{max}$ bounds, where $\calf$ consists of Dirichlet mixtures (with 10 components), as described above.
\item \textbf{Sample 100 priors for each hyperprior}. Sample $P(\cs)$ and $P(\ef\mid \cs)$ 100 times for each hyperpriors (that is, for each $\alpha$ setting).
\item \textbf{Sample the causal label uniformly}. If chose $X\rightarrow Y$ then let $\pxy = P(\cs)P(\ef\mid\cs)$. If chose $Y\rightarrow X$, let $\pxy = transpose[P(\cs)P(\ef\mid\cs)]$.
\item \textbf{Classify}. Use the LR classifier to classify $\pxy$'s causal direction and record ``error'' if the causal label disagrees with the classifier.
\end{compactenum}

\begin{figure}
\centering
\includegraphics[width=.5\textwidth]{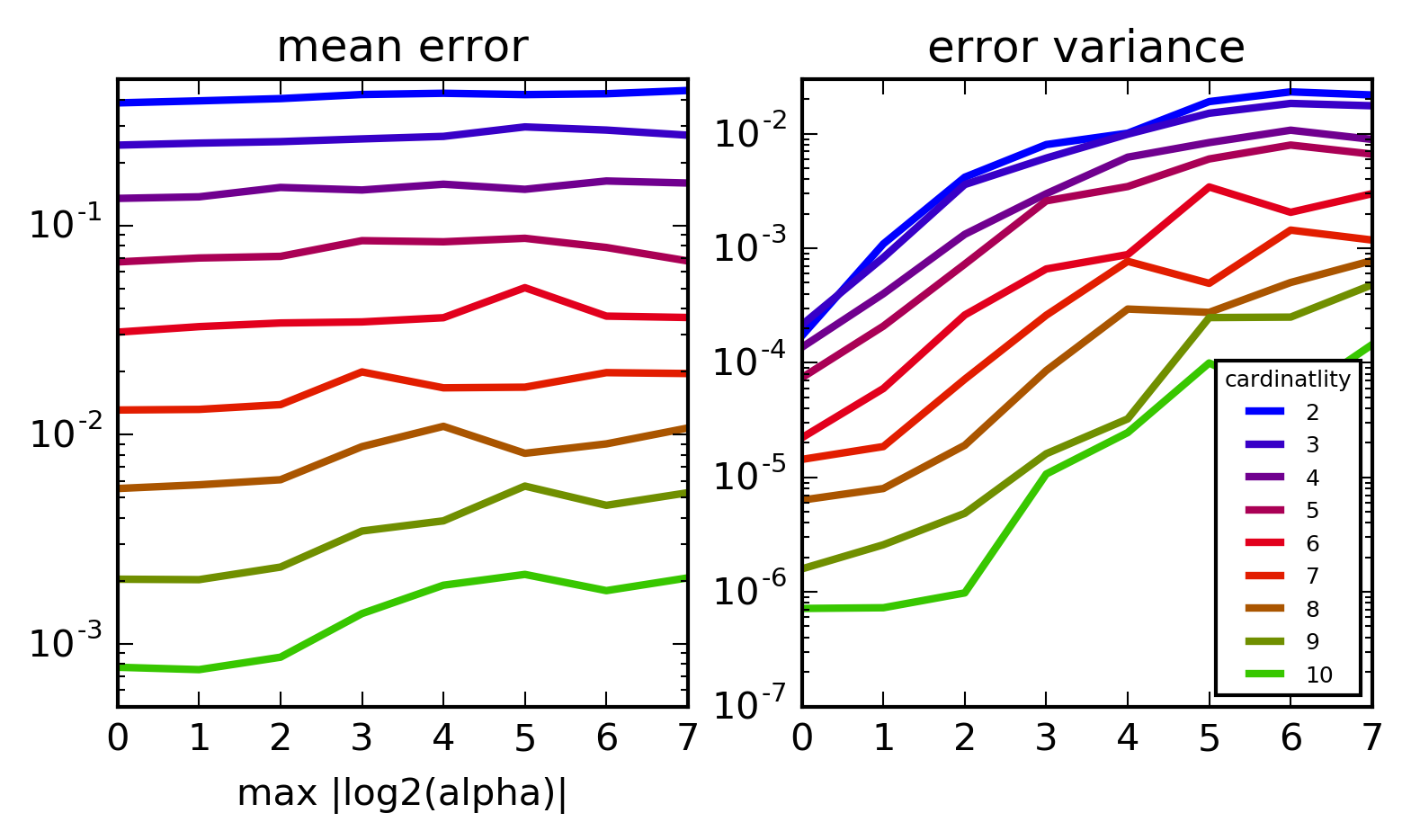}
\caption{Results of the direction-classification experiment. We varied cardinality of $X, Y$ as well as $\alpha_{max}$ of the mixture of Dirichlets $\calf$. For each setting, we sampled 100 $\pxy$ distributions according to our causal model and recorded the classification error of the simple $LR$ classifier. The results show that, as cardinality of $X$ and $Y$ grows, the $LR$ classifier's accuracy increases.}
\label{fig:dmm_results}
\end{figure}

\begin{figure}
\centering
\includegraphics[width=.5\textwidth]{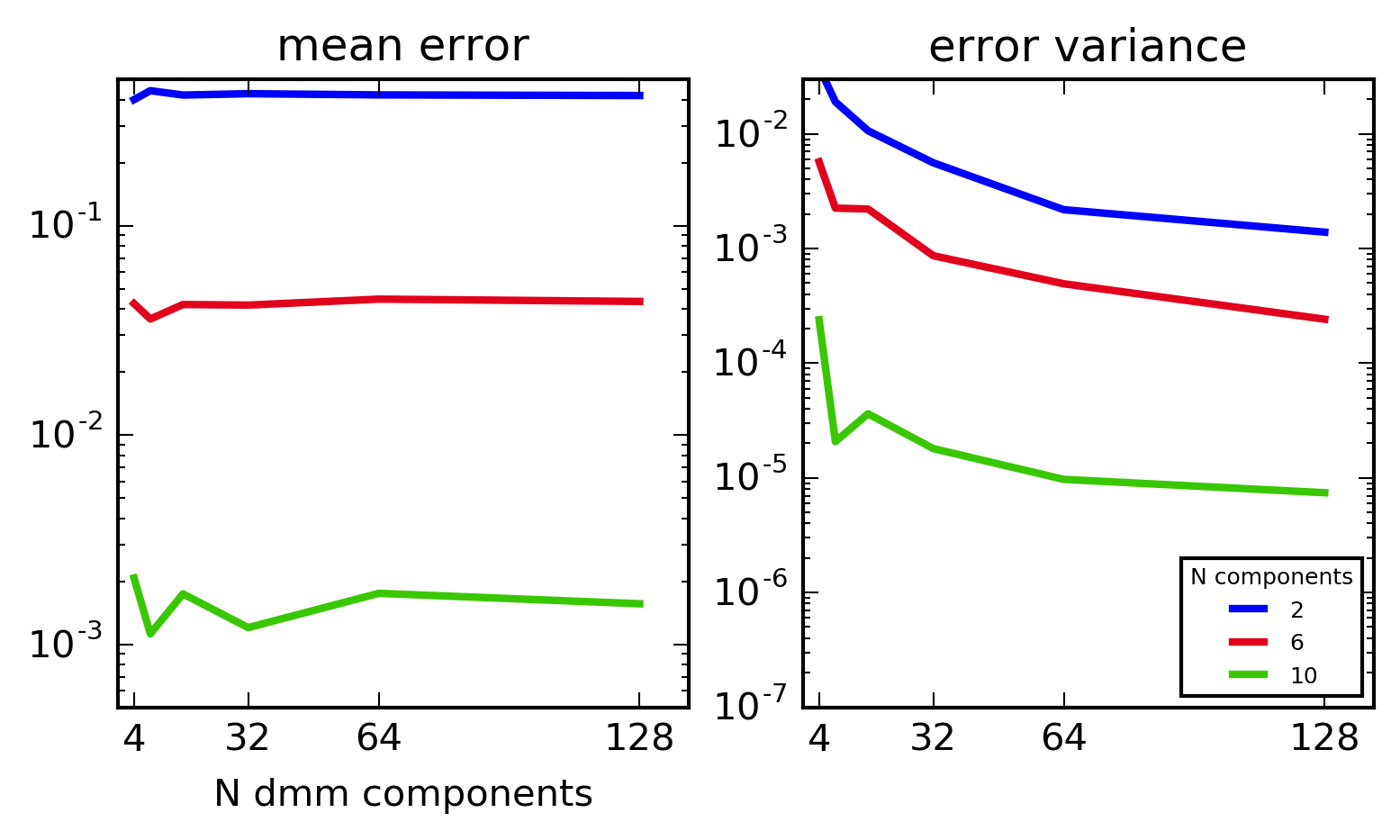}
\caption{Results of the direction-classification experiment when the number of Dirichlet mixture model hyperprior components varies. We fixed $\alpha$ to vary between $2^{-7}$ and $2^7$. The results show that the max-likelihood classifier that assumes the noninformative priors is not sensitive to the number of Dirichlet mixture components that the test data is sampled from.}
\label{fig:dmm_results_components}
\end{figure}

Figure~\ref{fig:dmm_results} shows the results. As the cardinality of the system grows, the LR classifier's decision boundary approximates the decision boundary for most Dirichlet mixtures. Another trend is that as $\alpha_{max}$ grows, the variance of the error grows, but there is only a small growing trend in the error itself. In addition, Fig.~\ref{fig:dmm_results_components} shows that the error does not increase as we allow more mixture components, up to 128 components, while holding $\alpha_{max}$ at the large value of 7. Thus, the LR classifier performs well even for extremely complex hyperpriors, at least on average.

\section{A Black-box Solution: Detecting Confounding}
\label{sec:confounding}
Consider now the question of whether $X \rightarrow Y$ or $X \rightarrow H \rightarrow Y$, where $H$ is a latent variable (a confounder). In this section we present a solution to this problem, under assumptions from Sec.~\ref{sec:assumptions}.

Unfortunately, deriving the optimal classifier for this case is difficult without additional assumptions on the latent $H$. Instead, we propose a black-box classifier. We created a dataset of distributions from both the direct-causal and confounded case, using the uninformative Dirichlet prior on either $P(X)$ and $P(Y\mid X)$ (the direct-causal case) or $P(H)$, $P(X\mid H)$ and $P(Y\mid H)$ in the confounded case. For each confounded distribution, we chose the cardinality of $H$, the hidden confounder, uniformly at random between 2 and 100. Next, we trained a neural network to classify the causal structure (Python code that reproduces the experiment is available at \url{vision.caltech.edu/~kchalupk/code.html}). We then checked how well this classifier performs as we vary the cardinality of the variables, and as we allow the true hyperprior to be a mixture of 10 Dirichlets, analogously to the experiment from Sec.~\ref{sec:analytical}.

\begin{figure}
\centering
\includegraphics[width=.5\textwidth]{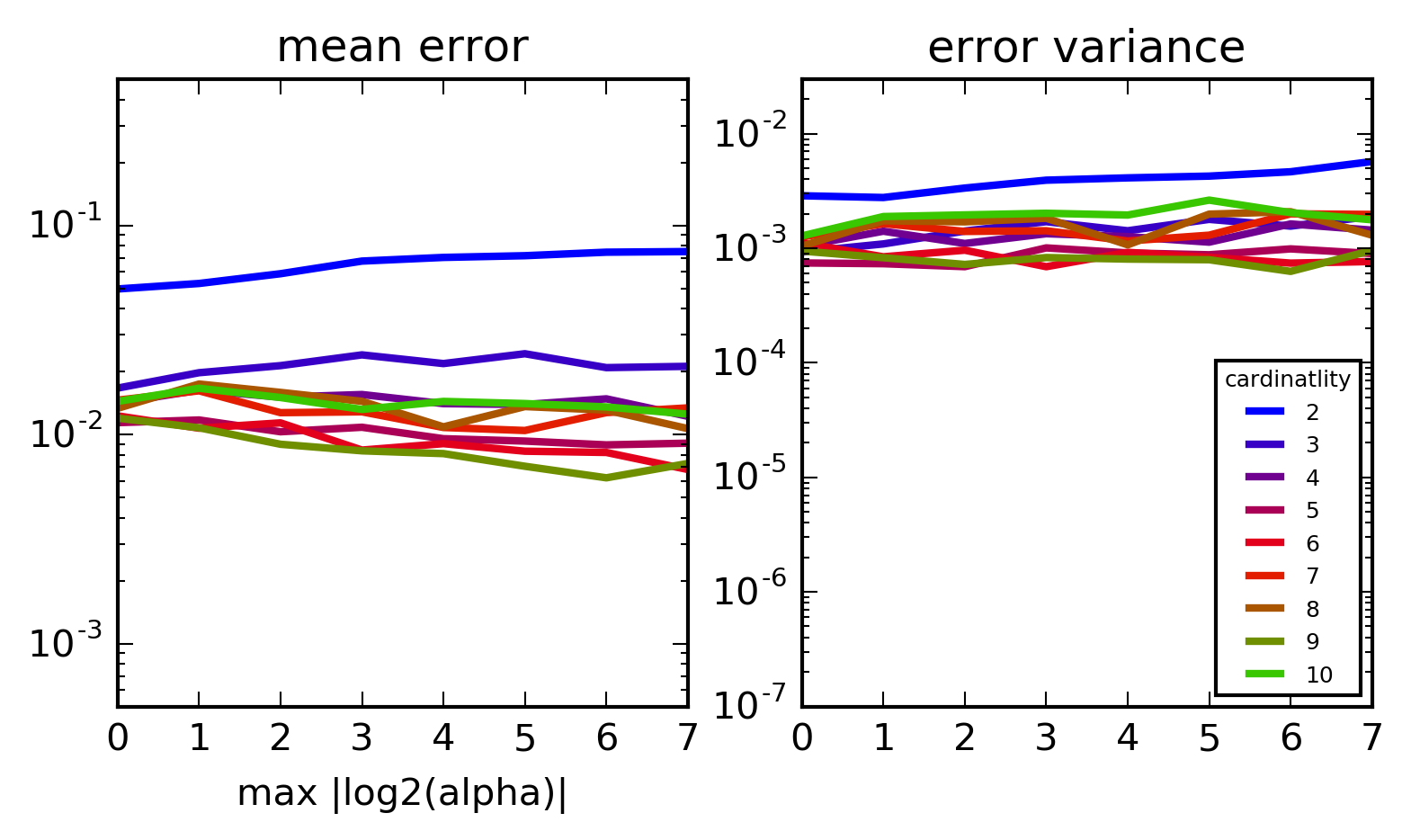}
\caption{Results of the black-box confounding detector. We varied cardinality of $X, Y$ as well as $\alpha_{max}$ of the mixture of Dirichlets $\calf$. For each setting, we sampled 1000 $\pxy$ distributions according to our causal model and recorded the classification error of a neural net classifier trained on noninformative Dirichlet hyperprior data. The results show that, as cardinality of $X$ and $Y$ grows, the $LR$ classifier's accuracy increases.}
\label{fig:nn_results}
\end{figure}

Fig.~\ref{fig:nn_results} shows the results. Note that the classification errors are much lower than for the ``deciding causal direction'' case. Both problems (deciding causal direction and detecting confounding) are in principle unidentifiable, but it appears the latter is inherently easier. The neural net classifier seems to be little bothered by growing $\alpha_{max}$. The largest source of error, for cardinality of $X$ and $Y$ larger than 3, seems to be neural network training rather than anything else.

\section{A Black-Box Solution to the General Problem}
Finally, we present a solution to the general causal discovery problem over the two variables $X$, $Y$: deciding between the six alternatives shown in Fig.~\ref{fig:causal_alternatives}. The idea is a natural extension of the black-box classifier from Sec.~\ref{sec:confounding}. We created a dataset containing all the six cases, sampled under the assumptions of Sec.~\ref{sec:assumptions}. We then trained a neural network on this dataset (the neural network architecture, as well as the details of the training procedure, are available in the accompanying Python code).

\begin{figure}
\centering
\includegraphics[width=.4\textwidth]{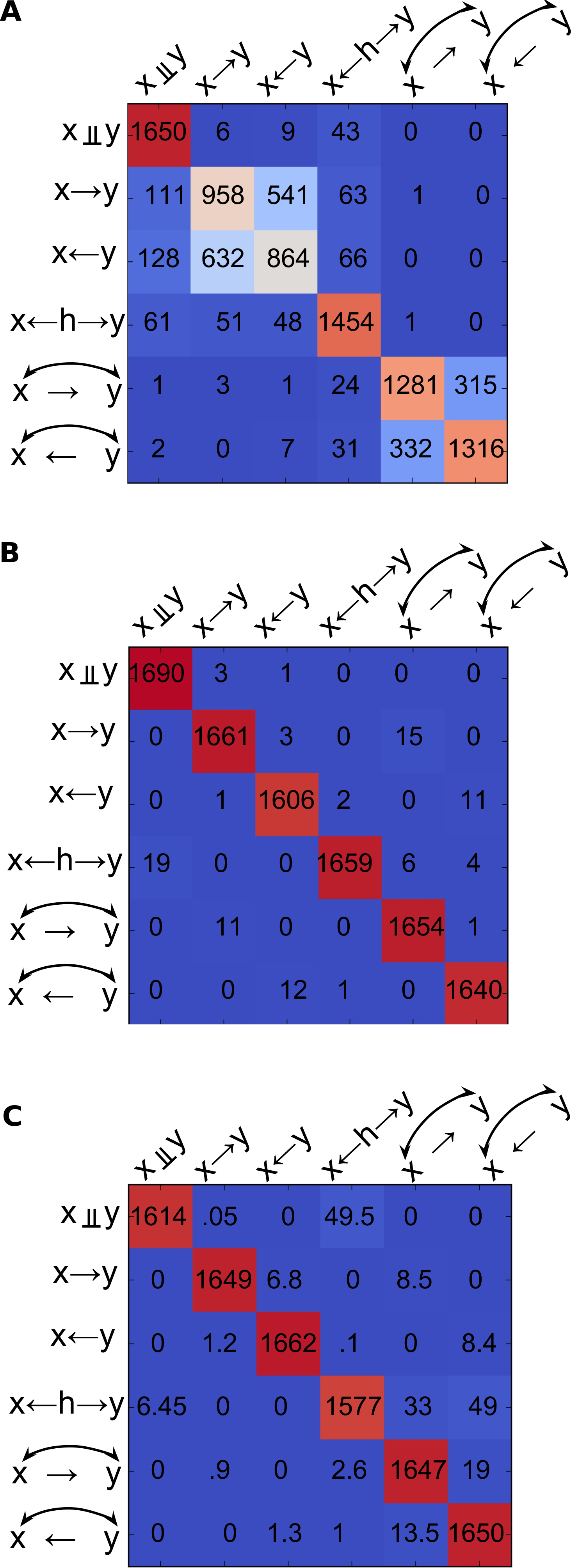}
\caption{Confusion matrices for the all-causal-classes classification task. The test set consists of distributions sampled from uniform hyperpriors -- that is, sampled from the same statistics as the training data (equivalent to $\alpha_{max}=0$ in previous sections). A) Results for $|X| = |Y| = 2$. Total number of errors=2477. B) Results for $|X|=|Y|=10$, total errors=85. C) Average results for $|X|=|Y|=10$, same classifier as in B) but test set sampled with non-uniform hyperpriors with $\alpha_{max}=7$ (see text). 201 errors on average. In each case, the test set contains 10000 distributions, with all the classes sampled with an equal chance.}
\label{fig:all_results}
\end{figure}

Figure~\ref{fig:all_results} shows the results of applying the classifier to distributions sampled from flat hyperpriors (that is, from a test set with statistics identical to the training set), for cardinalities $|X| = |Y| = 2$ and $|X|=|Y|=10$. As expected, the number of errors is much lower for the higher cardinality. For the cardinality of 2, the confusion matrix shows that the neural networks:
\begin{compactenum}
\item easily learn to classify independent vs dependent variables,
\item confuse the $X \rightarrow Y$ and $Y \rightarrow X$ cases, and
\item confuse the two ``directed-causal plus confounding'' cases (Fig.~\ref{fig:causal_alternatives}E,F).
\end{compactenum}
However, all these are insignificant issues when $|X|=10$, where the total error is 85 out of 10000 testpoints. For $|X|=2$, the error is 25.7\%. We remark again that the problem is not identifiable -- that is, there is no ``true causal class'' for any point in our training or test dataset. Each distribution \emph{could} arise from any of the possible five causal systems in which $X$ and $Y$ are not independent. The fact that the error nears 0 in the high-cardinality case indicates that the likelihoods under our assumptions grow very peaked as the cardinality grows. Thus, the \emph{optimal} decision can quite safely be called the \emph{true} decision. In addition, Fig.~\ref{fig:all_results}C shows the average confusion table for a hundred trials in which our classifier was applied to distributions over $X$ and $Y$ with cardinality 10, corresponding to all the possible six causal structures, but sampled from non-uniform hyperpriors with $\alpha_{max}=7$. The performance drop is not drastic compared to Fig.~\ref{fig:all_results}B.

\section{Discussion}
We developed a neural network that determines the causal structure that links two discrete variables. We allow for confounding between the two variables, but assumed acyclicity. The classifier takes as input a joint probability table $P_{XY}$ between the two variables and outputs the \emph{most likely} causal graph that corresponds to this joint. The possible causal graphs span the range shown in Fig.~\ref{fig:causal_alternatives} - from independence to confounding co-occurring with direct causation. We emphasize two limitations of the classifier:
\begin{compactenum}
\item Since the classifier makes a forced choice between the six acyclic alternatives, it will necessarily produce 100\% error on $P_{XY}$'s generated from cyclic systems.
\item Our goal was not, and can not be, to achieve 100\% accuracy. For example, error in Fig.~\ref{fig:all_results}A is about 25\%. However, this is not necessarily a ``bad'' result. Our considerations in Sec.~\ref{sec:assumptions}~and~\ref{sec:analytical} show that even when all our assumptions hold, the \emph{optimal} classifier has a non-zero error. 
\end{compactenum}

The latter is a consequence of the non-identifiability of the problem: it is not possible, in general, to identify the causal structure between two variables by looking at the joint distribution and without intervention. Our goal was to introduce a minimal set of assumptions that, while acknowledging the nonidentifiability, enable us to make useful inferences. 

We noted that as the cardinality of the variables raises, the task becomes more and more ``identifiable'' in the sense that, for each given $P_{XY}$, one out of the possible six causal graphs strongly dominates the others with respect to its likelihood. In this situation, the \emph{most likely} causal structure becomes essentially the only possible one, barring a small error, and the problem becomes \emph{practically} identifiable.

All of the above applies assuming that our generative model corresponds to reality. The assumptions, discussed in Sec.~\ref{sec:assumptions}, boil down to two ideas: 1) The world creates causes independently of causal mechanisms and 2) Causes are random variables whose distributions are sampled from flat Dirichlet hyperpriors. Causal mechanisms are conditional distributions of effects given causes, and are also sampled from flat Dirichlet hyperpriors. Whether these assumptions are realistic or not is an undecidable question. Nevertheless, through a series of simple experiments (Fig.~\ref{fig:dmm_results}, Fig.~\ref{fig:dmm_results_components}, Fig.~\ref{fig:nn_results}, Fig.~\ref{fig:all_results}) we showed that the assumption of flat hyperpriors is not essential -- our classifiers' average performance does not decrease significantly as we allow the hyperpriors to vary, although the variance of the performance grows. In future work, we will carefully analyze under what conditions the flat-hyperprior classifier performs well even if the hyperpriors are not flat. The current working hypothesis is that as long as the hyperprior on $P(cause)$ is the same as the hyperprior on $P(effect \mid cause)$, the classification performance doesn't change significantly \emph{on average},  but --as seen in our experiments -- it will have increased variance.

Shohei Shimizu explained our task (for the case of continuous variables) as: ``Under what circumstances and in what way can one determine causal structure based on data which is not obtained by controlled experiments but by passive observation only?''~\citep{Shimizu2006}. Our answer is, ``For high-cardinality discrete variables, it seems enough to assume independence of $P(cause)$ from $P(effect\mid cause)$, and train a neural network that learns the black-box mapping between observations and their causal generative mechanism.''

\bibliographystyle{plainnat}
\bibliography{bibliography}

\begin{thebibliography}{17}
\providecommand{\natexlab}[1]{#1}
\providecommand{\url}[1]{\texttt{#1}}
\expandafter\ifx\csname urlstyle\endcsname\relax
  \providecommand{\doi}[1]{doi: #1}\else
  \providecommand{\doi}{doi: \begingroup \urlstyle{rm}\Url}\fi

\bibitem[Chickering(2002{\natexlab{a}})]{Chickering2002}
David~Maxwell Chickering.
\newblock Learning equivalence classes of bayesian-network structures.
\newblock \emph{Journal of machine learning research}, 2\penalty0
  (Feb):\penalty0 445--498, 2002{\natexlab{a}}.

\bibitem[Chickering(2002{\natexlab{b}})]{Chickering2002b}
David~Maxwell Chickering.
\newblock Optimal structure identification with greedy search.
\newblock \emph{Journal of machine learning research}, 3\penalty0
  (Nov):\penalty0 507--554, 2002{\natexlab{b}}.

\bibitem[Geiger and Pearl(1988)]{gp1988}
D.~Geiger and J.~Pearl.
\newblock On the logic of causal models.
\newblock In \emph{Proceedings of UAI}, 1988.

\bibitem[Heckerman et~al.(1995)Heckerman, Geiger, and
  Chickering]{Heckerman1995}
David Heckerman, Dan Geiger, and David~M Chickering.
\newblock Learning bayesian networks: The combination of knowledge and
  statistical data.
\newblock \emph{Machine learning}, 20\penalty0 (3):\penalty0 197--243, 1995.

\bibitem[Hoyer et~al.(2009)Hoyer, Janzing, Mooij, Peters, and
  Sch{\"o}lkopf]{Hoyer2009}
Patrik~O Hoyer, Dominik Janzing, Joris~M Mooij, Jonas Peters, and Bernhard
  Sch{\"o}lkopf.
\newblock Nonlinear causal discovery with additive noise models.
\newblock In \emph{Advances in neural information processing systems}, pages
  689--696, 2009.

\bibitem[Hoyer et~al.(2008)Hoyer, Shimizu, Kerminen, and Palviainen]{hskp2008}
P.O. Hoyer, S.~Shimizu, A.J. Kerminen, and M.~Palviainen.
\newblock Estimation of causal effects using linear non-{G}aussian causal
  models with hidden variables.
\newblock \emph{International Journal of Approximate Reasoning}, 49:\penalty0
  362--378, 2008.

\bibitem[Janzing et~al.(2009)Janzing, Peters, Mooij, and
  Sch{\"o}lkopf]{Janzing2009}
Dominik Janzing, Jonas Peters, Joris Mooij, and Bernhard Sch{\"o}lkopf.
\newblock Identifying confounders using additive noise models.
\newblock In \emph{Proceedings of the Twenty-Fifth Conference on Uncertainty in
  Artificial Intelligence}, pages 249--257. AUAI Press, 2009.

\bibitem[Meek(1995)]{Meek1995}
C.~Meek.
\newblock {Strong completeness and faithfulness in Bayesian networks}.
\newblock In \emph{Eleventh Conference on Uncertainty in Artificial
  Intelligence}, pages 411--418, 1995.

\bibitem[Mooij et~al.(2014)Mooij, Peters, Janzing, Zscheischler, and
  Sch{\"o}lkopf]{Mooij2014}
Joris~M Mooij, Jonas Peters, Dominik Janzing, Jakob Zscheischler, and Bernhard
  Sch{\"o}lkopf.
\newblock Distinguishing cause from effect using observational data: methods
  and benchmarks.
\newblock \emph{arXiv preprint arXiv:1412.3773}, 2014.

\bibitem[Pearl(2000)]{Pearl2000}
J.~Pearl.
\newblock \emph{{Causality: Models, Reasoning and Inference}}.
\newblock Cambridge University Press, 2000.

\bibitem[Peters et~al.(2011)Peters, Janzing, and Scholkopf]{Peters2011}
Jonas Peters, Dominik Janzing, and Bernhard Scholkopf.
\newblock Causal inference on discrete data using additive noise models.
\newblock \emph{IEEE Transactions on Pattern Analysis and Machine
  Intelligence}, 33\penalty0 (12):\penalty0 2436--2450, 2011.

\bibitem[Reichenbach(1991)]{Reichenbach1991}
Hans Reichenbach.
\newblock \emph{The direction of time}, volume~65.
\newblock Univ of California Press, 1991.

\bibitem[Shimizu et~al.(2006)Shimizu, Hoyer, Hyv{\"a}rinen, and
  Kerminen]{Shimizu2006}
Shohei Shimizu, Patrik~O Hoyer, Aapo Hyv{\"a}rinen, and Antti Kerminen.
\newblock A linear non-gaussian acyclic model for causal discovery.
\newblock \emph{Journal of Machine Learning Research}, 7\penalty0
  (Oct):\penalty0 2003--2030, 2006.

\bibitem[Spirtes et~al.(2000)Spirtes, Glymour, and Scheines]{Spirtes2000}
P.~Spirtes, C.~N. Glymour, and R.~Scheines.
\newblock \emph{{Causation, prediction, and search}}.
\newblock Massachusetts Institute of Technology, 2nd ed. edition, 2000.

\bibitem[Sun et~al.(2006)Sun, Janzing, and Sch{\"o}lkopf]{Sun2006}
Xiaohai Sun, Dominik Janzing, and Bernhard Sch{\"o}lkopf.
\newblock Causal inference by choosing graphs with most plausible markov
  kernels.
\newblock In \emph{ISAIM}, 2006.

\bibitem[{Theano Development Team}(2016)]{Theano}
{Theano Development Team}.
\newblock {Theano: A {Python} framework for fast computation of mathematical
  expressions}.
\newblock \emph{arXiv e-prints}, abs/1605.02688, May 2016.
\newblock URL \url{http://arxiv.org/abs/1605.02688}.

\bibitem[Zhang and Hyv{\"a}rinen(2009)]{Zhang2009}
Kun Zhang and Aapo Hyv{\"a}rinen.
\newblock On the identifiability of the post-nonlinear causal model.
\newblock In \emph{Proceedings of the twenty-fifth conference on uncertainty in
  artificial intelligence}, pages 647--655. AUAI Press, 2009.

\end{thebibliography}

\end{document}